\documentclass[twocolumn,10pt]{article}

\usepackage[utf8]{inputenc}
\usepackage{xspace}
\usepackage[inline,shortlabels]{enumitem}
\usepackage[numbers,sort&compress]{natbib}
\usepackage[font=small,labelfont=bf]{caption}
\usepackage{graphicx}
\usepackage{amsmath}
\usepackage{fancyhdr}
\usepackage{verbatim}
\usepackage[hidelinks]{hyperref}
\usepackage{xcolor}
\usepackage{amsfonts}
\usepackage{libertine}
\usepackage{flushend}

\graphicspath{{images/}}

\newcommand{\projectname}{BusTr\xspace}
\DeclareMathOperator{\Relu}{ReLU}

\title{\projectname : Predicting Bus Travel Times from Real-Time Traffic}

\author{Richard Barnes \\ UC Berkeley \\ {rbar@berkeley.edu}
\and Senaka Buthpitiya \\ Google Research \\ {senaka@google.com}
\and James Cook \\ Google Research \\ {jcook@cs.berkeley.edu}
\and Alex Fabrikant \\ Google Research \\ {fabrikant@google.com}
\and Andrew Tomkins \\ Google Research \\ {tomkins@google.com}
\and Fangzhou Xu \\ Google Research \\ {fangzhoux@google.com}}

\date{}

\fancypagestyle{firstpagestyle}{
\lhead{\footnotesize{Cite as \textbf{Richard Barnes, Senaka Buthpitiya, James Cook, Alex Fabrikant, Andrew Tomkins, Fangzhou Xu (2020). ``BusTr: Predicting Bus Travel Times from Real-Time Traffic". 26th ACM SIGKDD Conference on Knowledge Discovery and Data Mining. doi: 10.1145/3394486.3403376}}}
\rhead{}
\cfoot{\thepage}
}

\begin{document}

% Enable page numbers
%\settopmatter{printfolios=true}
\maketitle
\thispagestyle{firstpagestyle}

\begin{abstract}
We present \projectname, a machine-learned model for translating road traffic forecasts into predictions of bus delays, used by Google Maps to serve the majority of the world's public transit systems where no official real-time bus tracking is provided. We demonstrate that our neural sequence model improves over DeepTTE, the state-of-the-art baseline, both in performance ($-30\%$ MAPE) and training stability. We also demonstrate significant generalization gains over simpler models, evaluated on longitudinal data to cope with a constantly evolving world.
\end{abstract}

\section{Introduction}
\label{sec:intro}
We present \projectname , a real-time delay forecasting system for public buses, which is used by Google Maps to expand the availability of real-time data for transit users around the world~\citep{ettblog}.

Public transit systems are vital to human mobility in our rapidly urbanizing world. World-wide, transit is the most common mode for trips after walking \citep{aguilera2014}. Public transit investments and availability continue to grow, driven especially by the many societal benefits of public transit over private transportation, from reduced congestion~\citep{Anderson2014}, to environmental impacts~\citep{Vincent2006,ipcc2014,Mendoza2019}, to social impacts~\citep{Pathak2017,Chetty2018}.

A modern global-scale mapping and navigation service thus needs to serve the needs of
transit users. What are these needs? Broadly, a transit user wants to know (1) what the transit system is \textit{supposed to do}: the system's routes, stops, and schedules, and (2) what the transit system \textit{is doing right now}: the current locations and delays of the transit trips, which often deviate significantly from published schedules~\citep{Wessel2017}. Of these two modalities, the real-time state is disproportionately important for the routine trips that dominate most people's transportation needs. Most transit users know by heart the routes connecting their home, work, and other frequent destinations, but they have a well-established need for information about real-time changes. Transit variability is a source of rider anxiety and a barrier to increasing
ridership~\citep{Brakewood2014,wob2016,Ferris2010,Chakrabarti2015,Zhang2008,Watkins2011}, and users place significant value on commute time reliability~\citep{Lam2001}.

Google Maps and other public transit apps are typically built on transit data distributed via the GTFS protocol~\citep{gtfsstatic} for static data, and its GTFS-Realtime extension~\citep{gtfsrt} for real-time tracking of public transit vehicle locations and delays. Ideally, every public transit agency would instrument its vehicles with networked real-time tracking hardware and provide a fresh, precise, and open feed of the location data. Anecdotally, many agencies are interested in such a system, but, as of 2020, the vast majority of the world's GTFS feeds with static transit data do not yet come with a matching real-time feed, due to a variety of operational constraints on the transit agencies' capabilities. Furthermore, even if an agency is able to reliably maintain tracking devices on its entire vehicle fleet, generating a useful real-time transit data feed requires live labeling of vehicles with transit metadata (via algorithmic approaches \citep{LiveTravel}, integration with dispatching solutions, or labor-intensive operator input). Any given agency can certainly overcome these barriers with a sufficient investment of capital and operating expenses, but here we aim for a solution to meet the needs of a global-scale transit tracking product.

An alternative to agency-driven solutions is to crowd-source the real-time location of transit vehicles~\citep{transitapp}, but this is infeasible to do with global-scale coverage while still fully protecting user privacy: plenty of transit trips will have too few users providing vehicle location data. Other crowd-sensing options hinge on activity recognition on mobile devices: inferring from a device's sensors what type of vehicle it's being transported on in real time. Distinguishing buses from other road vehicles via on-device sensors with usably high quality remains an open research question~\citep{guvensan2018}.

\subsection{Our approach: \projectname}

With \projectname , we pursue a different approach: we infer bus delays from a combination of real-time road traffic forecasts and contextual information about the transit and road systems learned from historical data. This focuses our attention on transit affected by road traffic: buses, rather than trains and subways.

Note that we specifically use real-time traffic to estimate \textit{delays}, or \textit{estimated travel times} (ETTs) between pairs of stops. A transit user typically seeks two figures in real-time: the ETD,  \textit{estimated time of departure} from their source stop, and the ETA, \textit{estimated time of arrival} to their destination stop, where $\mathrm{ETA}=\mathrm{ETD}+\mathrm{ETT}$. 
In the common case of journeys where bus headways, gaps between consecutive buses on the same line, are much shorter than typical trip times, we expect that ETTs dominate the user's information need. Estimating absolute ETDs and ETAs is infeasible without directly tracking the bus in real time, especially without optimistic assumptions about on-schedule departures from the stop of origin. 

Road traffic forecasts are obtained from crowd-sensed data, a well-studied approach~\citep{wan2016}. In our deployment, the road traffic forecasts come from Google Maps. Buses are not cars, though. Due to stops, schedule constraints, bus-specific road rules, and other dynamics of bus movement, bus delays are substantially different from car delays on the same roads~\citep{mcknight04}. \projectname\ combines real-time road traffic forecasts with contextual information about the transit system learned from historical data and the static features of the transit system, yielding $2.7\times$ overall error reduction over the baseline of using off-the-shelf road traffic forecasts directly (Sec.~\ref{sec:simplebase}).

To learn such a model, we need labeled examples of bus trips labeled with the incurred delays, combined with historical data about traffic on relevant roads. To learn about the peculiarities of local transit systems, road networks, and human movement dynamics, we need the training data to have as high coverage as possible in terms of space and time. In practice, such data is necessarily sparser and more heterogeneous than ideal, and can come from a mix of different sources, such as after-the-fact bus data provided by public transit agencies, user-contributed labels, road loop detectors, etc. To allow as many different data sources as possible, we optimize the system for training on a minimal set of features and strong generalization to areas and transit features never seen at training. For reproducibility, we focus our experiments here on training data from transit systems that do provide real-time transit data via GTFS-Realtime, but we heavily strip down the training data format and data density to allow the system to generalize to other settings, as detailed in Sec.~\ref{sec:datasets}.

The other features used by the model, detailed in Sec.~\ref{sec:model} are relatively spartan. While some prior work~\citep{Salvo2007,Julio2016} relies on detailed metadata such as bus lane locations and turn lanes, we expect that this data won't be available with high coverage, quality, and freshness at a global scale. Instead, we rely on our model to infer local features of the transit and road networks on various scales from the training data.

In Sec.~\ref{sec:experiments}, we measure the performance of \projectname\ on held-out data. We focus on comparisons against simple baselines, and against a state-of-the-art system described in~\citep{Wang2018}. We also demonstrate the importance of the features of the model and the training protocol with ablation tests, and show how our model generalizes to data not seen at training time, to adjust to a changing world.

\section{Related work}
\label{sec:related}
% While the literature on public transit prediction is expansive, only a few works focus on the problem of using traffic data to predict transit time. We review these here.
There is some existing literature on predicting bus travel times based on road traffic speeds, either measured using inductive loop detectors or inferred from bus speeds. We review some of this work here, then conclude by highlighting the differences between this work and our own.

\citet{Salvo2007} compare the performance of a multilayer perceptron (MLP) and a radial basis function (RBF) network in predicting the average speed of a bus over a segment of road. They find that the RBF performs better, but come to this conclusion by training on a very small dataset (112 points) drawn from a single bus line; generalization was tested by comparing against a second line. They break the bus's route into several segments and, for each, generate several features: traffic flow and capacity, whether or not there is a reserved bus lane, number of intersections with and without traffic lights, number of bus stops, whether there is illegal parking or free parking present, the number of inlets and outlets to the segment, the number of pedestrian crossings, and whether or not ``commercial activities" are present. Details on the final network structure are omitted. The paper does not describe how traffic data was measured, only that it was provided by the Public Transport Company of Palermo. On unseen data, their RBF had a MAPE of 9\% and their MLP had a MAPE of 34\%.

\citet{Mazloumi2011} note that while previous approaches focus on predicting average bus travel times, the variability in travel time is often neglected. Accordingly, they train two fully-connected neural networks---one to predict average time and the other its variance---each with a single hidden layer on an 1,800 point dataset for a four segment (five stop) route. Traffic variance is assumed to be normal about the mean, though they note that there can be long tails in delays. Training features include: traffic speed within each segment, measured using inductive loop detectors and averaged over a variable time window; schedule adherence (delay relative to the timetable); and temporal variables (day of week, time of day, month of year). They find that weather does not influence their predictive accuracy, possibly due to the lower number of training examples, so they omit this from their model. A neural network with a single hidden layer is used. After training networks of various sizes with Bayesian regularization, networks with 2--3 nodes turn out to provide the best accuracy. They find that traffic information adds little additional value beyond temporal variables alone.

\citet{Sun2016} predict arrival times at various bus stops by calculating the delay versus a scheduled time. They distinguish between cases where the predicted time of arrival is in the near versus far future. Far future delays are found by dividing the data into seven groups by day of week, then within each group using k-means to cluster delay data according to the delay and the time of day to produce between 2 and 5 clusters. For arrival times in the near future, a two-stage Kalman filter is used. The first stage uses the bus's reported location to develop an estimate of its true position. The second stage uses the position to estimate the delay of the bus on its current segment. While the first stage of the filter is updated on a per-bus basis, the second stage updates each segment using information from possibly many buses whose routes overlap. Information is drawn from GTFS static and real-time data as well as historical bus timing data; using traffic flow is listed as future work. The model was deployed in Nashville, TN, USA and reduced hour-ahead arrival prediction errors by an average of 25\% and 15-minute errors by 47\%.

\citet{Julio2016} compare the performance of multi-layer perceptrons, SVMs, and Bayes Nets on predicting bus travel speeds from traffic conditions (the bus's real-time location is used as a proxy for traffic), finding that MLPs performed best. To do so, they discretize each bus's trajectory into a space-time grid where each cell represents about 400\,m distance and 15--30 minutes of time. It is unclear whether these cells aggregate statistics from multiple bus lines or only multiple buses on a single line. From this information they extract eight potential features: real-time and historic speeds for the incoming, current, and outgoing cell over the previous ten minutes, historical speeds for the current cell at the moment to be predicted, and a binary variable indicating whether the cell contains a bus-only lane. Forward selection narrows the features to only the real-time speeds of the downstream and current cell, as well as the historic speed of the current cell. This information was fed to an MLP with two hidden layers of size 6 and 5 (structure obtained via trial-and-error). Predictive accuracy declined for times with high congestion, so k-means was used to multiplex models across possible traffic conditions. MAPE ranged from 14--22\%.

\citet{Dhivyabharathi2019} use real-time bus location information as a proxy for traffic with the aim of predicting travel times over each segment of a trip. They note that their data has a log-normal distribution and build two predictors around this: a seasonal AR model with possibly non-stationary effects and a linear non-stationary AR model. The seasonal model performs better with a MAPE of 17--19\%, as tested on a single bus route. They compare this against an MLP of unspecified structure, trained on the travel times of recent trips through a segment, with a MAPE of 20--24\% on the same route. Notably, the MLP has less feature diversity than in other works and is trained with Levenberg--Marquardt back propagation rather than the Bayesian regularization approach preferred by other authors.

\begin{comment}
Bus arrival time prediction using artificial neural network model
R. Jeong, R. Rilett, 2004
From related work doc:
* Listing it because Julio et al listed it.
* Talks about “congestion”, but it looks like “congestion” is estimated using historical and realtime schedule adherance; doesn’t look like they actually have a separate source of road traffic data.
* Two sections of a single bus route in Houston.

Improved iterative prediction for multiple stop arrival time using a support vector machine
Chang-Jiang Zheng, Yi-Hua Zhang, Xue-Jun Feng, 2011
From related work doc:
* Listing it because Julio et al listed it.
* Doesn’t seem to directly use traffic data: “However, traffic conditions are complicated and difficult to measure, and thus this research uses bus speed on the road link for estimating traffic conditions of the links”
\end{comment}

\citet{jeong2004bus} and \citet{zheng2012improved} also use bus location traces as a proxy for traffic data when modelling bus travel times.

The DeepTTE system presented in~\citet{Wang2018} predicts transit times between locations. Their deep neural model first converts raw latitude-longitude pairs from GPS trackers to 16-dimensional vectors. A convolution is run across each time series and the results concatenated with embedded metadata features (such as the day of the week and the weather). This is then passed through a two-high stacked LSTM. Two things now happen. (1)~The LSTM time series outputs are passed through densely connected layers to give per-segment timing predictions. (2)~The LSTM time series outputs are combined with the metadata again in an attention layer. The result is again concatenated and passed through a series of residual fully-connected layers to give a prediction for the travel time across all segments. The per-segment and overall predictions are jointly used to train the model for which they report a MAPE of 11.89\% in Chengdu and 10.92\% in Beijing. In Sec.~\ref{sec:wangbaseline} we use this model as a baseline for its state-of-the-art performance and its deep network structure, comparably modern to \projectname .

% None of the above authors were able to provide source code or replication materials for their work.

Including the broader literature of predicting bus arrival times, non-neural methods are the dominant approach and perform well \citep{Reich2019}, but shallow perceptrons of only 1--3 layers show similar or better performance while potentially providing superior generalization versus deeper nets~\citep{Mei2007,Mazloumi2011}. Despite this, more recent work has shown good performance with deep nets~\citep{Treethidtaphat2017,Heghedus2017}, recurrent nets~\citep{Heghedus2019}, and attention (MAPE 14.8\%)~\citep{Sun2019}. Several authors have found it advantageous to cluster historical travel information and use this as part of a multiplexed prediction approach~\citep{Julio2016,Xu2017,Sun2016}. This may offer advantages over MLPs because MLPs may have difficulty accounting for disruptions or out-of-band events~\citep{Reich2019}. \citet{Reich2019} note that the lack of standard benchmarks and open source code make inter-comparison difficult.

\begin{comment}
Studies not included:

Bus arrival time prediction at bus stop with multiple routes
https://drive.google.com/file/d/1bbPwB1_TZjRvQo_vdtKpg6GoPOqhwcsB/view?usp=sharing
Bin Yu, William H.K. Lam, Mei Lam Tam, 2011
Notes
Listing it here because Julio et al listed it.
Doesn’t appear to use traffic data. (Based on the list following “The factors considered in this study can be illustrated as follows.”)
Lists a bunch of past papers about predicting bus travel times.

Bus Arrival Time Prediction Using Support Vector Machines
Yu Bin, Yang Zhongzhen, Yao Baozhen, 2007
Notes
Listing it here because Julio at al listed it.
Contribution: “This article presents support vector machines (SVM), a new neural network algorithm, to predict bus arrival time.”
Doesn’t seem to use traffic (based on reading the abstract): “Segment, the travel time of current segment, and the latest travel time of next segment are taken as three input features”
\end{comment}

\textbf{Our approach differs from previous work in several key respects.}
(a)~Our model is developed with generalization in mind. Our model should provide reasonable estimates of traffic-bus relations both for new routes in cities for which we have training data, as well as for cities in which we have no training data. The existing literature (with the exception of \citep{Salvo2007}) focuses on improving predictions for known bus routes without regard to generalization.
(b)~Our model uses a restricted feature set. \citet{Salvo2007} notes that features noting ``bus only" lanes, commercial activities, and illegal parking all add significant predictive power to their model; however, acquiring such information globally is difficult. Instead, the spatial elements of our model allow it to infer the existence of these features when they are present by learning both local and regional characteristics of the space a bus route passes through.
(c)~Our model is trained with a much larger amount of data. While previous authors have performed their analysis around single bus routes, we consider our model's performance on a planet-scale dataset. This allows us to avoid having to incorporate strong priors such as log-normality~\citep{Dhivyabharathi2019}.
(d)~Our model makes inferences from real-time traffic data. Previous work used real-time bus locations as a proxy for traffic information, thus limiting generalization, or traffic loop sensors, which are sparse and usually confined to major roadways.

\section{Datasets}
\label{sec:datasets}
To forecast a travel time, \projectname\ needs two points on a bus route to delimit the trip; road traffic speed info for the relevant streets and times; and contextual data for the trip: the identity of the bus route, the roads involved, and time-of-week.

At training time, we need golden data: a clean, validated, integrated dataset with durations of specific bus trip segments, aligned with road traffic speeds at the relevant time. Here, we focus on training on data provided by GTFS-Realtime feeds via ``Vehicle Position'' reports, which specify the live locations of transit vehicles. Inference in this setting can actually add a delay forecast to a fresh Vehicle Position to provide an absolute ETA estimate, but this is not our primary focus. Instead, we aim to build and evaluate a model that can estimate delays for bus lines where there is no GTFS-realtime data, just a sporadic flow of offline observations of bus timings  from a variety of sources, which will likely not have full coverage of bus lines, roads, and/or timings, and may also substantially vary in frequency, regularity, and precision of bus location observations.

To work in such a setting, we first represent our input data as training examples that are just pairs of timed trip endpoints, without finer-grained information on the timing at points in-between. Similarly to text mining, we \textit{shingle} an input trajectory, here a sequence of GTFS-RT vehicle positions, into possibly-overlapping examples with several heuristic constraints:

\begin{itemize}
\item We avoid shingle endpoints at or near stops. Although user queries will typically pertain to bus delays between pairs of stops, there is extra uncertainty inherent in a vehicle position reported at a stop: we cannot tell whether the bus just arrived at the stop or is just departing. These two states represent a noticeable difference in a bus's progress through a trip. Since vehicle positions may be reported imprecisely, we also exclude reports that are near to a stop.
Instead, we use vehicle positions reported at other points along the bus trip polyline, which various data sources including GTFS will often have.
\item For each input bus trajectory, we sample a minimum shingle length uniformly from $[1, 5]$ km, and pick shingles of at least this length. This approximates a common range of user trip lengths, avoids shingles that are short enough that their observed duration is likely subsumed by noise in the endpoint location, and, by sampling from a wide range, forces the model to not overfit on typical shingle length.
\item 
The start times of shingles extracted from the same input trajectory are spaced at least 30 seconds apart, to limit data redundancy from very densely reported trajectories
\item We remove outlier shingles during which a gap between consecutive trajectory reports exceeded corpus-specific values ($5$\,min or $3$\,km). Shingles with unlikely reported average speeds (outside $[0.7, 140]$\,km/h) are also excluded.
\end{itemize}

Our shingling intentionally does not attempt to resample or interpolate between the points in raw location reports because we expect relevant bus motion to be non-uniform, especially when a pair of location reports spans a stop, a long red light, or a localized traffic snarl.

Shingling can confound simple protocols for holding out data, since adjacent shingles from one trajectory are not independent. In our experiments, we separate training, validation, and test sets by calendar weeks, using a separate 7-day span of data for each. This also gives us a way to measure generalization of the model as the world evolves over time, addressed further in Sec.~\ref{sec:generalization}.

Road traffic forecasts are obtained from Google Maps, on a per road segment basis. A single road segment is, roughly, a stretch of road between two adjacent turns. Since we train offline, we train using the traffic speed estimates that were available at the time the bus traversed a segment, estimated by the underlying road traffic system from the best available combination of aggregate real-time data and historical inferences. At inference time, the model can rely on the underlying system to provide forward forecasts of traffic per road segment, with the expectation that training on "cleaner" data without forecasting errors will not bias the model.

\section{Model}
\label{sec:model}
\begin{figure}
\includegraphics[width=\columnwidth]{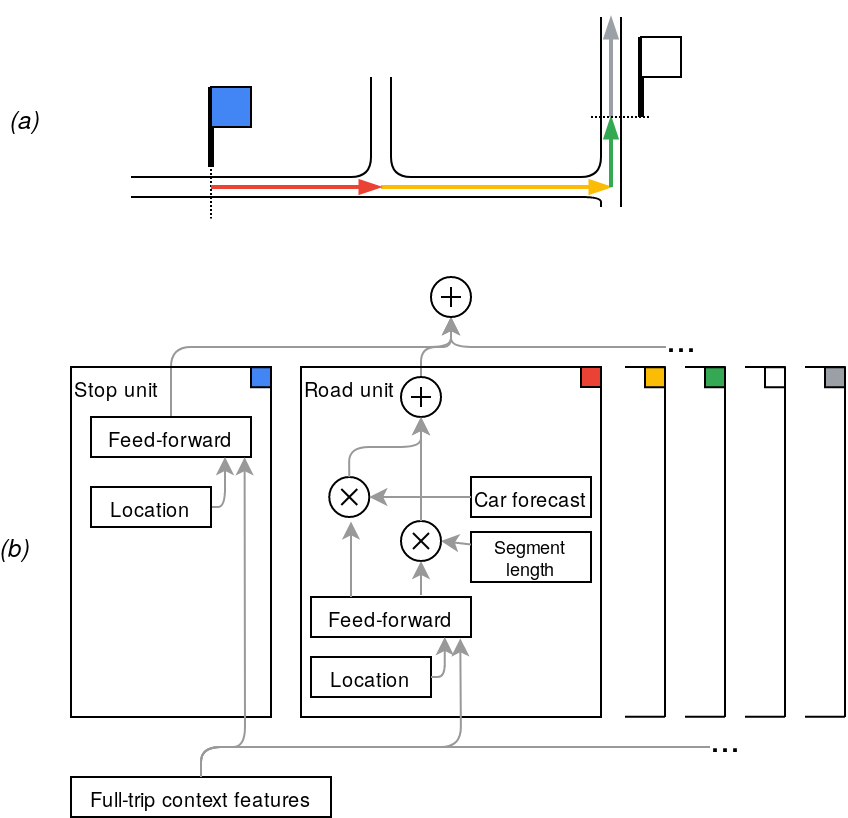}
\caption{Model Overview. (a) An example trajectory including several stops and segments. Note that segments are shorter than the distance between two stops. (b) A depiction of the model's overall structure. Information about the trajectory, as well as each stop and segment on the trajectory, is used to produce independent estimates of the travel time across each stop and segment. Finally, these estimates are summed to estimate the travel time of the route as a whole. \label{fig:mo}}
\end{figure}

The job of the model is to predict how long a bus will take to travel along a given interval of its route, based on traffic conditions and the current time.

The model separately predicts the time taken to traverse each road segment and service each stop, both denoted $t_q$ where $q$ is a road segment or stop.
These predictions are summed to produce a prediction $\hat{T}$ of the total trip duration $T$ as the final output:
$$\hat{T} = \sum_{q \in Q} \hat{t}_q$$
where $Q$ is the set of road segments and stops in the trip interval, and $\hat{t}_q$ is the model's prediction of $t_q$.

\subsection{Structure of an example}

The structure of the model reflects the structure of the examples given to the model, so we start by describing those. An example consists of:

\begin{itemize}
\item \emph{A trip interval.}
Each example is built around an interval of a bus trip. At inference time, this interval is typically between an arbitrary pair of stops. At training time, the interval is an arbitrary shingle with endpoints not aligned to stops, as described in Sec.~\ref{sec:datasets}.
\item \emph{A sequence of \emph{quanta} constituting that interval.} The trip interval is divided into a sequence of road segments and stops, hereafter ``quanta''. For example, the trip interval shown in \autoref{fig:mo}(a) becomes the following sequence: the blue starting stop; three road segments (red, yellow, and green, each a separate quantum); the next stop (white); and part of the next road segment (gray).
\item \emph{Traffic speeds and other per-quantum features.} Each quantum has an associated feature vector, described in Sec.~\ref{sec:quantumFeatures}.
\item \emph{Full-trip context features.} The trip as a whole has an associated feature vector, described in Sec.~\ref{sec:contextFeatures}.
\end{itemize}

At training time, we require a \emph{prediction target}: the time duration $T$ the bus took to traverse the trip interval.

\begin{table}
\newcommand{\SEL}[1]{\textcolor{red}{#1}}
    \centering
    \begin{tabular}{l|l|l}
% Hidden layer size
Learning rate & $\eta$ & $0.01, 0.03, \SEL{0.1}, 0.3$ \\
Decay rate & $\delta$ & $0.97$ every $1000$ steps \\
Hidden layer size & $k$ & $16, \SEL{32}$ \\
S2 cell levels & $L$ & [15, 12.5, 4.5] \\
Ablation rates & $p_L$ & [0.2, 0.1, 0.1] \\
$L_1$ regularizer base & $b$ & $1.25$ \\
$L_1$ regularizer & $\alpha$ & 0.001, 0.01, \SEL{0.1}, 1.0 \\
Feature selection threshold & $\varepsilon$ & $0.1$ \\[6pt]
Embedding dimensions: & & \\
Hour of day & $d_h$ & $0, \SEL{2}, 4$ \\
Day of week & $d_w$ & $0, \SEL{2}$ \\
Spatial & $d_s$ & $0, \SEL{4}, 8$ \\
Route & $d_r$ & $0, \SEL{2}, 4$
    \end{tabular}
    \caption{Model hyperparameters. Vizier was used to select the red ones over the black ones where given; others were set manually.}
    \label{tab:hyperparams}
\end{table}

\subsection{Full-trip context features}
\label{sec:contextFeatures}

Each example carries two context features describing the sequence as a whole:

\begin{itemize}
\item A $d_r$-dimensional embedding of the \textit{bus route identifier}. Contrary to GTFS practice, we use ``route'' to refer to the exact ordered sequence of stops and the public route identity. So, ``bus 5 northbound'', ``bus 5 southbound'', and ``2am run of bus 5 skipping stop X'' are treated as three distinct routes.
\item The \textit{time} when the bus is at the start of the trip interval. In training data, this is the observed time; at inference time, this is typically the current wall time.
\end{itemize}

Similarly to DeepTTE~\citep{Wang2018}, we represent time by discretizing and embedding it, expecting to encourage our network to learn a more nuanced representation of time than it would likely do so by operating on numerical values alone. Time of week is discretized to two values: a day of the week and a half-hour slice of the day. These are embedded separately in $\mathbb{R}^{d_w}$ and $\mathbb{R}^{d_h}$, respectively. While we initialize other embeddings randomly, we've observed that the model eventually arranges times of day roughly in a contiguous cycle in the target space. We shortcut this by initializing the first two dimensions of the embedding to put the time on a circle, so that the model immediately starts out with a notion of ``similar'' times of day. In trials with $d_h > 2$, the other dimensions are initialized with Gaussian noise.

Note that we consciously use time as a context feature rather than a per-quantum feature. In our configuration, both training shingles and user trips are typically so short that they rarely span meaningfully different times of day (e.g.\ ``rush hour'' vs ``late evening''). In ad hoc experiments, we did not see a measurable quality lift from estimating a separate per-quantum ``time as of traversal'' feature.

We train on data on the scale of weeks, and thus do not attempt to directly capture seasonality effects beyond what is captured by the real-time traffic data implicitly.

\subsection{Per-quantum features}

Each quantum has associated features. Both stops and road segments get an embedding of the quantum's location, described in \ref{sec:spaceRepr} below. Stops get no other features. Road segments get two more features, described in \ref{sec:quantumFeatures} below.

\subsubsection{Representing locations}
\label{sec:spaceRepr}

Each trajectory is a sequence of narrowly defined locations: a stop or a road segment (typically shorter than 100\,m). Yet, we aim to capture spatial variation in bus behavior on many scales of locality, hoping to balance the model's needs to respond to both narrowly local phenomena such as bus pullouts or bus lanes and regional phenomena like national traffic laws, with the need to generalize well to specific locations not seen verbatim in the training data.

For each quantum, we take a representative point --- a road segment's endpoint or a stop's GTFS-reported point location --- and discretize it using cell identifiers from the S2 Geometry Library (s2geometry.io), which provides a discrete global grid hierarchy (see~\citep{Sahr2003,Barnes2019} for a review of alternatives).
We start with a ``level 15'' S2 cell, a roughly square quadrilateral of area about 0.08\,km$^2$. We also go ``up'' the hierarchy to coarser cells containing that one. We somewhat arbitrarily use ``level 12.5'' and ``level 4.5'' cells (i.e.\ adjacent pairs of level 13 and 5 S2 cells). These quadrilaterals are roughly $2:1$ rectangles of area 2.5\,km$^2$ and $1.3\times 10^6$\,km$^2$. These three levels correspond, roughly, to a small neighborhood, a town, and a metropolitan area.  Each cell identifier is separately embedded into $\mathbb{R}^{d_s}$, and these embeddings, for each quantum separately, are combined together with an unweighted sum. This redundant representation is regularized (see Secs.~\ref{sec:training}) to encourage the model to learn heavier weights for the coarser cells where possible, with the more numerous coarse cells only getting embedded with a substantial norm when there's a need to represent something more local.

Spatial representation is shared across stop and road features, in hopes that some salient unobserved features such as crowdedness patterns may contribute to both computations.

\subsubsection{Additional features for road segments}
\label{sec:quantumFeatures}

For road segments, we use the segment's length as a feature. In cases where the stop or trip interval endpoint lands in the middle of a segment, we use only the length actually traversed.

We also obtain a road traffic speed estimate from Google Maps, as described in Sec.~\ref{sec:datasets}. To get the forecast, we must first decide on a target time for that forecast. Ideally, this would be the time at which the bus will reach that road segment, but at inference time, we only know the start time of the trip.
To solve this, we estimate when the bus will reach the segment by starting with the trip's start time and crudely extrapolating based on car travel time along previous legs. For consistency, this method is used to retrieve historical traffic speeds at training time, as well.

An alternative to car travel time would be to iteratively use the model's own predictions, but this would limit parallelizability at inference time, and introduce additional model complexity. In early ad hoc experiments, we did not observe substantial quality gains from more sophisticated approaches to propagating absolute timestamps through the sequence of quanta.

\subsection{Model structure}

The model consists of one unit for each quantum $q$ that makes up the trip interval. Each unit produces a prediction $\hat{t}_q$ of how long the bus will spend traversing that segment or waiting at that stop. The model outputs the sum $\hat{T} = \sum_{q \in Q} \hat{t}_q$ of these per-unit predictions as an estimate of the overall trip duration, as shown in \autoref{fig:mo}(b).

Each unit in the model has a single fully-connected hidden layer of width $k$, with ReLU activation, with a single set of hidden layer weights shared across all the units of both types.

\subsubsection{Stop units}

To predict time spent at a stop, the unit simply connects the hidden layer to all context and per-quantum features, and to a single output node responsible for the prediction.

\subsubsection{Road segment units}

For road segments, two of the features are handled specially: the road traffic speed forecast $s$ and the distance $d$ travelled on the segment. The hidden layer is connected directly to all of the other context and per-quantum features, and to two intermediate outputs $\alpha$ and $\beta$ which are used as coefficients in a sum:
\begin{equation*}
\textrm{Segment duration} = \alpha \frac {d} {s} + \beta d
\end{equation*}
(Here, $d/s$ is an estimate of the time a car would take to traverse the segment.)
This formula effectively lets the unit learn a simple linear mixture of ``car-dependent'' and ``car-independent'' impacts on the time taken to traverse the segment.

\subsubsection{Post-processing}

The duration produced by each unit is further clipped, by replacing negative durations with 0. The model's output for the example is the sum of the clipped per-unit values.

\subsubsection{Summary}

For a stop $q$, the time estimate is computed as
$$ \hat{t}_q = \Relu\left( W_2 \Relu \left(W_1 \begin{pmatrix}\ell_q \\ c\end{pmatrix} \right) \right)$$
where $W_2$ are the weights connecting the output to the hidden layer, $W_1$ is the matrix of weights connecting the hidden layer to the location embedding,  $\ell_q$ is the location embedding, $c$ is the concatenated full-trip context features, and $\Relu(x) = \max\{0, x\}$.

For a road segment $q$, the time estimate is computed as
\begin{align*}
\hat{t}_q =& \Relu(\alpha_q d_q / s_q + \beta_q d_q) \\
\begin{pmatrix}\alpha_q \\ \beta_q\end{pmatrix} =&
    W_2 \Relu\left( W_1 \begin{pmatrix}\ell_q \\ c\end{pmatrix}\right)
\end{align*}

\section{Training}
\label{sec:training}
\subsection{Spatial input ablation}

To push the model to generalize better without losing training data coverage, we \textit{simulate novelty} in training data by \textit{spatial input ablation}, i.e. by removing fine-grained spatial and route features. For each training example, we pick an ablation level L: ``remove route id and spatial cell features at level $L$ and finer for all quanta'' with probability $p_L$ for each $L$, otherwise ``keep all features''. This allows the model to experience seeing examples that are ``far from what I know about'', for various degrees of ``far''. This is applied to the whole trajectory to avoid leaking spatial information across quanta. On the other hand, for another example on the same route or in the same area, the spatial input ablation policy is sampled independently, so the model can still find a good embedding for the local features. We intend for this to push the larger spatial embedding weights ``up'' the spatial hierarchy unless there's something special about the fine-grained location.

\subsection{Feature selection}

In an additional effort to move the spatial embedding weights ``up'' to coarser cells, we train the model twice. In the first pass, we apply an $L_1$ regularizer to the \textit{average} embedding in each layer, weighted with $\alpha b^L$, where $\alpha$ and $b$ are hyperparameters we tune for and $L$ is the S2 cell level (per Table \ref{tab:hyperparams}). The exponentially higher penalty aims to get near-zero embeddings for most fine cells that don't need them. At the end of the first pass, we select only the spatial features that got embeddings with norm above $\varepsilon = 0.1$, and then re-train the model from scratch, with no $L_1$ regularization, while discarding any spatial features that were not selected in the first round.

This has the additional benefit of significant model size reduction. In 20 trials, the size of the embedded vocabulary, dominated by level-15 spatial cells, is shrunk by an average of $2.2\times$ ($\sigma = 0.060$).

\subsection{Training protocol}
We train our model using the Adam optimizer~\citep{kingma2014adam} with MSE loss, using 100,000 training steps and 200 training examples per step. We evaluate the model on 100,000 examples sampled from the validation set every 500 training steps, and select the checkpoint with the lowest MAPE on the validation set. We use Adam since most of our model's parameters are for embeddings of very sparse features, which Adam is designed for.

\subsection{Hyperparameter tuning}
\label{sec:hyperTuning}
We tuned the model's hyper-parameters with the results shown in Table~\ref{tab:hyperparams} using Vizier~\citep{Golovin2017,Solnik2017} with 64 trials, at 100,000 training steps each.

We further validated the 100K-step duration at Vizier's reported optimal parameters with 20 trials at 25K, 50K, 100K, 200K, and 400K steps. 100K-step training performed modestly, but statistically significantly, better on MAPE over a held-out test set (Table~\ref{tab:stepcount}).
\begin{table}
    \centering
    \begin{tabular}{r|c|l}
        Steps & Test MAPE (stdev) & p-value \\
\hline
25K & 13.886 (0.084) & $p\ll 10^{-10}$ \\
50K & 13.480 (0.065) & $p\ll 10^{-10}$ \\
100K & 13.240 (0.045) &  $-$ \\
200K & 13.337 (0.063) & $p< 8\times 10^{-6}$ \\
400K & 13.377 (0.087) & $p< 9\times 10^{-7}$
    \end{tabular}
    \caption{Test MAPE by training step count and p-value for 100K being optimal, with one tailed t-test with $4\times$ Bonferroni correction}
    \label{tab:stepcount}
\end{table}

\section{Experiments}
\label{sec:experiments}
We adopt per-shingle MAPE (mean absolute percentage error) as the target metric. A review of the ETA prediction literature~\citep{Reich2019} notes that inconsistencies in reporting standards prevent the inter-comparison of approaches. MAPE, they report, is the most common metric used by the studies reviewed (13 of 40 studies), and is thus our choice here, too.

Except where otherwise stated, all the experiments train a model 20 times and evaluate its performance on 100,000 examples sampled from the test data set. The test data comes from a week of calendar data that is not used during training or validation.

We show the mean and standard deviation of per-run test MAPEs. Statistical significance of improvements are evaluated with one-tailed t-tests, with the p-values shown.

\subsection{Simple baselines}
\label{sec:simplebase}

The zero-th order approximation to our problem is to just use car traffic to estimate bus delays. Anecdotally, users of many bus services without real-time data use car mapping services just like this. This approach, tested on 20 trials of 100k examples each from our test dataset produces mean MAPE 35.616 (stdev 0.060).

% TODO include visual from blog post?

Another natural baseline is a linear regression of the trajectory-scale features, without context, using just three per-trajectory numerical features: number of stops, distance traversed, and car traffic time estimates. With 20 trials of linear regression tested on disjoint data slices from our training week, evaluated on 100k disjoint slices of the test week, the per-trial mean MAPE is 22.944 (stdev 0.089).

The model's inference latency was fast enough in absolute terms that we did not rigorously compare computational performance gaps against the compute performance of simpler models. In one deployment, 90th percentile computation latency for a TensorFlow implementation of BusTr over a natural query distribution required per-query compute time below 38.1 msec, over $700\times$ faster than the typical 30-second reporting interval of GTFS-Realtime feeds.

\subsection{Comparison against DeepTTE}
\label{sec:wangbaseline}

As a state of the art baseline, we implemented DeepTTE~\citep{Wang2018} to run in our setting using our full set of features. We made two adjustments. First, we omitted the model components operating on intra-trajectory points, since we expect those to not be available in our setting. Second, for a fair comparison, we tuned hyperparameters for both DeepTTE and \projectname\ with Vizier using 10K steps of training for each, since training DeepTTE, a substantially more complex model, for 100K steps proved prohibitively slow.

For our model, the Vizier-optimized parameters at 10K steps only differed from Table~\ref{tab:hyperparams} by changing the learning rate settings to $\eta=0.03$ and $\delta=0.9$. For DeepTTE's contextual features, we used time of day, route ID, and spatial features at the same three S2 levels as \projectname. Vizier chose embedding dimensions of 2 for time, 2 for route ID, and 4 for each spatial level separately. Vizier gave $k=5$ filter kernel size; $c=32$ for the geo-conv layer size; 128 as the LSTM size; and set the learning rate at 0.01.

\begin{figure}
\includegraphics[width=\columnwidth]{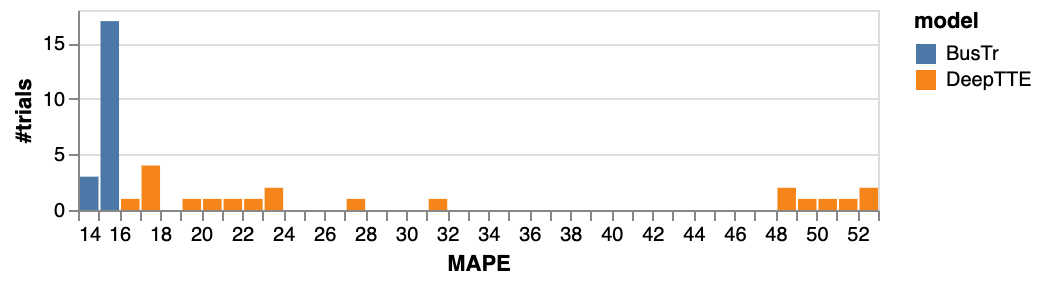}
\caption{MAPEs for BusTr vs DeepTTE from 20 training trials. \label{fig:deeptte-mape-hist}}
\end{figure}

We then evaluated both models on 20 independent training runs using the final model at step 10K on one week of data, and tested the performance on a held-out week of data. In addition to significantly slower training, we observed that DeepTTE, under the same conditions produces both significantly lower MAPE on average, and substantially worse convergence: in 37\% of trials, DeepTTE MAPEs never converged, staying significantly above 40\% (Fig.~\ref{fig:deeptte-mape-hist}). The mean MAPE performance of DeepTTE was statistically significantly worse than \projectname (Table~\ref{tab:deeptte}), even if we manually discarded the runs that did not converge as outliers.

\begin{table}
    \centering\footnotesize
    \begin{tabular}{l|l|l}
     & Test MAPE (stdev) & p-value \\
     \hline
\projectname & 15.164 (0.143) & $-$ \\
DeepTTE & 31.466 (14.368) & $p<8\times 10^{-6}$ \\
\begin{minipage}[c][6ex]{27mm}\vspace{0em}DeepTTE (excluding
MAPEs above 40\%)\end{minipage} & 21.242 (4.200) & $p<3\times 10^{-7}$ \\
\end{tabular}
\caption{\projectname\ vs DeepTTE, tuned for 10K-step training. We substantially outperform DeepTTE, even if we discard the runs where DeepTTE MAPEs don't converge. One-tailed t-test p-values are for losses compared to \projectname.}
    \label{tab:deeptte}
\end{table}

%%% rijard's experiment notes moved to richard-experiment-notes.txt

\subsection{Feature ablation}
\label{sec:feature-abl}

We now consider ablating several of the model's features one at a time to evaluate their contribution to the model's performance, as summarized in Table~\ref{tab:feature-abl}.

\begin{table}
    \centering\footnotesize
    \begin{tabular}{l|c|l}
Variant & Test MAPE (stdev) & p-value \\ \hline
Full model & 13.240 (0.045) &  $-$ \\
Road traffic ablated & 15.443 (0.139) & $p\ll 10^{-10}$ \\
Route IDs ablated & 13.602 (0.064) & $p\ll 10^{-10}$ \\
\begin{minipage}[c][6ex]{30mm}Route IDs and level-15 cells ablated\end{minipage} & 14.923 (0.074) & $p\ll 10^{-10}$ \\
\begin{minipage}[c][6ex]{30mm}Route IDs and all spatial cells ablated\end{minipage} & 22.190 (0.116) & $p\ll 10^{-10}$ \\
Time of week ablated & 13.865 (0.062) & $p\ll 10^{-10}$ \\
\begin{minipage}[c][6ex]{30mm}Numerical signals as generic hidden inputs\end{minipage} &
13.459 (0.107) & $p<4\times 10^{-10}$
    \end{tabular}
    \caption{Feature ablations, with one-tailed t-test p-values for losses compared to the full model, over $n=20$ runs}
    \label{tab:feature-abl}
\end{table}

We first evaluate the \textit{importance of real-time data} for the forecasts. With traffic data absent, the model remains free to make predictions informed by the current time and the location of the trip, but is capturing nothing about what makes today's behavior any different from historical data. This degrades the model's MAPE by $+16\%$.

We next consider fine-grained spatial features. We disable the route ID embedding --- is the fine spatial location of the trajectory sufficient without knowing which of the several bus routes running through the area is being queried? The MAPE here degrades by $+3\%$. We conjecture the modest loss is likely due to the few places where different bus routes are timed substantially differently, such as with asymmetric bus lanes, or express and local buses operating differently on the same road segments. Ablating spatial information further, to remove both the route ID and the finest (level 15) spatial cell, degrades the performance further, by $+13\%$, reflecting the importance of local structure to the model. Once we ablate route IDs and spatial information entirely, thus removing geographical context even at the metro and country level, the errors skyrocket by $+69\%$. Even with all the spatial features dropped, we still perform statistically significantly ($p\ll 10^{-10}$) above the linear baseline in Sec.~\ref{sec:simplebase}, but by a much thinner margin, just 4\%.

Adding back all the spatial information and ablating time signals (hour of day and day of week), we incur a $+5\%$ MAPE loss versus the full model. Note that hour of week also figures into both the observed real-time traffic, and into historical inferences made by the underlying road traffic forecaster, so this likely underestimates the true impact of temporal context.

Lastly, we consider the possibility of undoing the final linear layer in the model's architecture, instead allowing the model to learn a more arbitrary function of the observed speed and distance features, which also produces a small but statistically significant MAPE loss ($+2\%$).

\subsection{Generalization}
\label{sec:generalization}

Several of the features of our model and training protocol are especially designed to promote generalization. It turns out that removing these features actually produces modest quality improvements compared to the full model when tested in our default set up.

To measure the trade-off behind generalization features, we use a natural experiment: testing on data describing routes and locations that appeared in our GTFS data over a 2-month period and weren't available at training time. In particular, we compare results from testing on three distinct test datasets:

\begin{enumerate}
    \item ``1 week away - full'': A full test week of held-out data from trips during a time window 1 week away from the training data, as in Sec.~\ref{sec:feature-abl} above.
    \item A held-out test week 9 weeks away from training data, restricted to:
    \begin{enumerate}
        \item ``New routes over 9 weeks'' - Trip shingles on route IDs that were never seen at training time.
        \item ``New areas over 9 weeks'' - Trip shingles that at least once pass through a level-12.5 S2 cell that was never seen at training time.
    \end{enumerate}
\end{enumerate}

The ``new route ID'' case can capture new bus routes created in the world, changes to which stops a route visits, and new GTFS feed providers. The ``new L=12.5 cell'' case is likelier to capture whole neighborhoods that weren't previously served by buses, or weren't described with a GTFS feed available to us.

In these settings, we test ablating these generalization-oriented features:

\begin{itemize}
    \item Removing feature selection, to instead train with a single 100K-step round with no spatial regularization.
    \item Disabling spatial input ablation (SIA) at training time
    \item Disabling coarse S2 cells, leaving only level-15 features
    \item Disabling both spatial input ablation and S2 cells coarser than level 15 in the spatial hierarchy
\end{itemize}

The last ``double'' feature ablation accounts for the fact that just removing coarse cells makes spatial input ablation a much weaker proposition, since in the ablated examples, this would leave the model with no spatial context at all. Interestingly, the generalization losses are milder here than either feature in isolation, which we find unexpected.

\begin{table*}[t]
    \centering
    \begin{tabular}{l|ll|ll|ll}
Model &  \multicolumn{2}{c|}{1 week away - full} &  \multicolumn{2}{c|}{New routes over 9 weeks} &
\multicolumn{2}{c}{New areas over 9 weeks} \\
\hline \hline
Full model              & 13.240 (0.045) & $-$     & 15.495 (0.103) & $-$             & 19.581 (0.352) & $-$ \\
No feature selection    & 13.167 (0.058) & ${}^*$  & 16.074 (0.197) & $p\ll 10^{-10}$ & 22.667 (0.549) & $p\ll 10^{-10}$ \\
No SIA                  & 13.140 (0.048) & ${}^*$  & 15.997 (0.224) & $p< 10^{-10}$   & 21.810 (0.846) & $p\ll 10^{-10}$ \\
No coarse cells         & 13.258 (0.078) & $p>0.2$ & 16.685 (0.317) & $p\ll 10^{-10}$ & 23.604 (0.551) & $p\ll 10^{-10}$ \\
No SIA, no coarse cells & 13.052 (0.061) & ${}^*$  & 15.632 (0.227) & $p<0.011$       & 21.511 (1.220) & $p< 4\times 10^{-8}$
\end{tabular}
    \caption{Effect of generalization features on test data soon after training, and on novel data over 9-week span. Test MAPE with standard deviation in parentheses. One-tailed t-test p-values given where the full model's mean improves over the ablation (n=20 trials);
    ${}^*$ - trials where the system under test outperformed the full model.}
    \label{tab:gen}
\end{table*}

Table~\ref{tab:gen} summarizes the generalization experiments. We believe that foregoing the minor quality wins seen in the ``1 week away'' full test is worthwhile given the substantial generalization gains on novel data in both the ``new route'' and ``new area'' test sets.

By focusing on real changes to the ecosystem over 2 months, we provide a measurement of practically-relevant generalization.
An alternative experiment would be to instead use a synthetic model for holding out test data to simulate novelty. For instance, we can try applying spatial input ablation to the test data as well. Unsurprisingly, this gives an advantage to a model that's also trained with spatial input ablation: $4\%$ improvement in MAPE, statistically significant at $p\ll 10^{-10}$. However, this may well just speak to the synthetic assumptions made during training being better aligned to the synthetic assumptions during test, so we do not consider this as a separate strong argument to support our model's generalization.

\section{Conclusion}
\label{sec:conclusion}
We have described a new model, \projectname, for predicting how long it will take public transit buses to travel between points on their routes based on contextual features such as location and time as well as estimates of current traffic conditions. Our model demonstrates excellent generalization to test data that differs both spatially and temporally from the training examples we use, allowing our model to cope gracefully with the ever-changing world.

Our model outperforms not only simple predictors, but also DeepTTE, the previous state of the art. This is remarkable given the relative simplicity of our design. Our work shows that judicious feature selection and design choices, coupled with sufficient training data, can give superior results, in terms of both prediction accuracy and training cost, versus more complex designs.

Uncertainty regarding transit times for public transit buses is a barrier to increasing transit ridership; our work is another step in the direction of reducing this uncertainty.

\section*{Acknowledgements}
The authors thank Cayden Meyer for directing us toward this problem space; Da-Cheng Juan for his ML modeling insights; and Neha Arora, Anthony Bertuca, Matt Deeds, Julian Gibbons, Reuben Kan, Ivan Kuznetsov, Oliver Lange, David Lattimore, Thierry Le Boulengé, Ramesh Nagarajan, Marc Nunkesser, Anatoli Plotnikov, Ivan Volosyuk, and the greater Google Transit and Road Traffic teams for support, helpful discussions, and assistance with bringing this system to the world at large. We are also indebted to our partner agencies for providing the GTFS transit data feeds the system is trained on.

\footnotesize
\bibliography{refs.bib}

\begin{thebibliography}{43}
\providecommand{\natexlab}[1]{#1}
\providecommand{\url}[1]{\texttt{#1}}
\expandafter\ifx\csname urlstyle\endcsname\relax
  \providecommand{\doi}[1]{doi: #1}\else
  \providecommand{\doi}{doi: \begingroup \urlstyle{rm}\Url}\fi

\bibitem[Aguil{\'e}ra and Gr{\'e}bert(2014)]{aguilera2014}
Anne Aguil{\'e}ra and Jean Gr{\'e}bert.
\newblock Passenger transport mode share in cities: exploration of actual and
  future trends with a worldwide survey.
\newblock \emph{International Journal of Automotive Technology and Management},
  14\penalty0 (3-4):\penalty0 203--216, 2014.

\bibitem[Anderson(2014)]{Anderson2014}
Michael~L Anderson.
\newblock Subways, strikes, and slowdowns: The impacts of public transit on
  traffic congestion.
\newblock \emph{American Economic Review}, 104\penalty0 (9):\penalty0 2763--96,
  2014.

\bibitem[Barnes(2019)]{Barnes2019}
Richard Barnes.
\newblock Optimal orientations of discrete global grids and the poles of
  inaccessibility.
\newblock \emph{International Journal of Digital Earth}, 0\penalty0
  (0):\penalty0 1--14, 2019.

\bibitem[Brakewood et~al.(2014)Brakewood, Barbeau, and Watkins]{Brakewood2014}
Candace Brakewood, Sean Barbeau, and Kari Watkins.
\newblock An experiment evaluating the impacts of real-time transit information
  on bus riders in {T}ampa, {F}lorida.
\newblock \emph{Transportation Research Part A: Policy and Practice},
  69:\penalty0 409--422, 2014.

\bibitem[Chakrabarti and Giuliano(2015)]{Chakrabarti2015}
Sandip Chakrabarti and Genevieve Giuliano.
\newblock Does service reliability determine transit patronage? insights from
  the {Los} {Angeles} {Metro} bus system.
\newblock \emph{Transport Policy}, 42:\penalty0 12 -- 20, 2015.
\newblock ISSN 0967-070X.
\newblock URL
  \url{http://www.sciencedirect.com/science/article/pii/S0967070X15300068}.

\bibitem[Chen et~al.(2007)Chen, Yaw, Chien, and Liu]{Mei2007}
Mei Chen, Jason Yaw, Steven~I. Chien, and Xiaobo Liu.
\newblock Using automatic passenger counter data in bus arrival time
  prediction.
\newblock \emph{Journal of Advanced Transportation}, 41\penalty0 (3):\penalty0
  267--283, 2007.
\newblock URL
  \url{https://onlinelibrary.wiley.com/doi/abs/10.1002/atr.5670410304}.

\bibitem[Chetty and Hendren(2018)]{Chetty2018}
Raj Chetty and Nathaniel Hendren.
\newblock The impacts of neighborhoods on intergenerational mobility {I}:
  Childhood exposure effects.
\newblock \emph{The Quarterly Journal of Economics}, 133\penalty0 (3):\penalty0
  1107--1162, 2018.

\bibitem[Dhivyabharathi et~al.(2019)Dhivyabharathi, Kumar, Achar, and
  Vanajakshi]{Dhivyabharathi2019}
B.~Dhivyabharathi, B.~Anil Kumar, Avinash Achar, and Lelitha Vanajakshi.
\newblock Bus travel time prediction: A lognormal auto-regressive ({AR})
  modeling approach.
\newblock \emph{arXiv: 1904.03444}, 2019.

\bibitem[Fabrikant(2019)]{ettblog}
Alex Fabrikant.
\newblock Predicting bus delays with machine learning.
\newblock Google {AI} Blog, 2019.
\newblock URL
  \url{https://ai.googleblog.com/2019/06/predicting-bus-delays-with-machine.html}.

\bibitem[Ferris et~al.(2010)Ferris, Watkins, and Borning]{Ferris2010}
Brian Ferris, Kari Watkins, and Alan Borning.
\newblock {OneBusAway}: results from providing real-time arrival information
  for public transit.
\newblock In \emph{Proceedings of the SIGCHI Conference on Human Factors in
  Computing Systems}, pages 1807--1816. ACM, 2010.

\bibitem[Golovin et~al.(2017)Golovin, Solnik, Moitra, Kochanski, Karro, and
  Sculley]{Golovin2017}
Daniel Golovin, Benjamin Solnik, Subhodeep Moitra, Greg Kochanski, John Karro,
  and D.~Sculley.
\newblock Google {Vizier}: {A} {Service} for {Black}-{Box} {Optimization}.
\newblock In \emph{Proceedings of the 23rd {ACM} {SIGKDD} {International}
  {Conference} on {Knowledge} {Discovery} and {Data} {Mining} - {KDD} '17},
  pages 1487--1495, Halifax, NS, Canada, 2017. ACM Press.
\newblock ISBN 978-1-4503-4887-4.

\bibitem[GTFS({\natexlab{a}})]{gtfsrt}
GTFS.
\newblock {GTFS} {R}ealtime {S}pecification.
\newblock \url{https://developers.google.com/transit/gtfs-realtime/reference/},
  2020{\natexlab{a}}.

\bibitem[GTFS({\natexlab{b}})]{gtfsstatic}
GTFS.
\newblock {GTFS} static overview.
\newblock \url{https://developers.google.com/transit/gtfs}, 2020{\natexlab{b}}.

\bibitem[Guvensan et~al.(2018)Guvensan, Dusun, Can, and Turkmen]{guvensan2018}
M.~Amac Guvensan, Burak Dusun, Baris Can, and H.~Irem Turkmen.
\newblock A novel segment-based approach for improving classification
  performance of transport mode detection.
\newblock \emph{Sensors}, 18\penalty0 (1), 2018.

\bibitem[Heghedus(2017)]{Heghedus2017}
Cristina Heghedus.
\newblock {PhD} {Forum}: {Forecasting} {Public} {Transit} {Using} {Neural}
  {Network} {Models}.
\newblock In \emph{2017 {IEEE} {International} {Conference} on {Smart}
  {Computing} ({SMARTCOMP})}, pages 1--2, Hong Kong, China, May 2017. IEEE.
\newblock ISBN 978-1-5090-6517-2.

\bibitem[Heghedus et~al.(2019)Heghedus, Chakravorty, and Rong]{Heghedus2019}
Cristina Heghedus, Antorweep Chakravorty, and Chunming Rong.
\newblock Neural {Network} {Frameworks}. {Comparison} on {Public}
  {Transportation} {Prediction}.
\newblock In \emph{2019 {IEEE} {International} {Parallel} and {Distributed}
  {Processing} {Symposium} {Workshops} ({IPDPSW})}, pages 842--849, Rio de
  Janeiro, Brazil, May 2019. IEEE.
\newblock ISBN 978-1-72813-510-6.

\bibitem[{IPCC}(2014)]{ipcc2014}
{IPCC}.
\newblock \emph{{Climate Change 2014: Mitigation of Climate Change}}.
\newblock Cambridge University Press, 2014.
\newblock ISBN 978-1-107-05821-7.

\bibitem[Jeong and Rilett(2004)]{jeong2004bus}
Ranhee Jeong and R~Rilett.
\newblock Bus arrival time prediction using artificial neural network model.
\newblock In \emph{Proceedings. The 7th International IEEE Conference on
  Intelligent Transportation Systems (IEEE Cat. No. 04TH8749)}, pages 988--993.
  IEEE, 2004.

\bibitem[Julio et~al.(2016)Julio, Giesen, and Lizana]{Julio2016}
Nikolas Julio, Ricardo Giesen, and Pedro Lizana.
\newblock Real-time prediction of bus travel speeds using traffic shockwaves
  and machine learning algorithms.
\newblock \emph{Research in Transportation Economics}, 59:\penalty0 250 -- 257,
  2016.
\newblock ISSN 0739-8859.
\newblock Competition and Ownership in Land Passenger Transport (selected
  papers from the Thredbo 14 conference).

\bibitem[Kingma and Ba(2014)]{kingma2014adam}
Diederik~P. Kingma and Jimmy Ba.
\newblock Adam: A method for stochastic optimization, 2014.

\bibitem[Lam and Small(2001)]{Lam2001}
Terence~C. Lam and Kenneth~A. Small.
\newblock The value of time and reliability: measurement from a value pricing
  experiment.
\newblock \emph{Transportation Research Part E: Logistics and Transportation
  Review}, 37\penalty0 (2):\penalty0 231 -- 251, 2001.
\newblock ISSN 1366-5545.
\newblock Advances in the Valuation of Travel Time Savings.

\bibitem[Mazloumi et~al.(2011)Mazloumi, Rose, Currie, and Sarvi]{Mazloumi2011}
Ehsan Mazloumi, Geoff Rose, Graham Currie, and Majid Sarvi.
\newblock An integrated framework to predict bus travel time and its
  variability using traffic flow data.
\newblock \emph{Journal of Intelligent Transportation Systems}, 15\penalty0
  (2):\penalty0 75--90, 2011.

\bibitem[McKnight et~al.(2004)McKnight, Levinson, Ozbay, Kamga, and
  Paaswell]{mcknight04}
Claire McKnight, Herbert Levinson, Kaan Ozbay, Camille Kamga, and Robert
  Paaswell.
\newblock Impact of traffic congestion on bus travel time in northern new
  jersey.
\newblock \emph{Transportation Research Record}, 1884:\penalty0 27--35, 01
  2004.

\bibitem[Mendoza et~al.(2019)Mendoza, Buchert, and Lin]{Mendoza2019}
Daniel~L Mendoza, Martin~P Buchert, and John~C Lin.
\newblock Modeling net effects of transit operations on vehicle miles traveled,
  fuel consumption, carbon dioxide, and criteria air pollutant emissions in a
  mid-size {US} metro area: findings from {Salt} {Lake} {City}, {UT}.
\newblock \emph{Environmental Research Communications}, 1\penalty0
  (9):\penalty0 091002, Sep 2019.

\bibitem[{Osang} et~al.(2019){Osang}, {Cook}, {Fabrikant}, and
  {Gruteser}]{LiveTravel}
Georg {Osang}, James {Cook}, Alex {Fabrikant}, and Marco {Gruteser}.
\newblock Livetravel: Real-time matching of transit vehicle trajectories to
  transit routes at scale.
\newblock In \emph{Proceedings of 2019 IEEE ITSC}, pages 2244--2251, 2019.

\bibitem[Pathak et~al.(2017)Pathak, Wyczalkowski, and Huang]{Pathak2017}
Rahul Pathak, Christopher~K. Wyczalkowski, and Xi~Huang.
\newblock Public transit access and the changing spatial distribution of
  poverty.
\newblock \emph{Regional Science and Urban Economics}, 66:\penalty0 198 -- 212,
  2017.
\newblock ISSN 0166-0462.

\bibitem[Reich et~al.(2019)Reich, Budka, Robbins, and Hulbert]{Reich2019}
Thilo Reich, Marcin Budka, Derek Robbins, and David Hulbert.
\newblock Survey of {ETA} prediction methods in public transport networks.
\newblock \emph{arXiv: 1904.05037}, 2019.

\bibitem[Sahr et~al.(2003)Sahr, White, and Kimerling]{Sahr2003}
Kevin Sahr, Denis White, and A.~Jon Kimerling.
\newblock Geodesic discrete global grid systems.
\newblock \emph{Cartography and Geographic Information Science}, 30\penalty0
  (2):\penalty0 121--134, 2003.

\bibitem[Salvo et~al.(2007)Salvo, Amato, and Zito]{Salvo2007}
G.~Salvo, G.~Amato, and Pietro Zito.
\newblock Bus speed estimation by neural networks to improve the automatic
  fleet management.
\newblock \emph{European Transport}, 37:\penalty0 93--104, 2007.

\bibitem[Solnik et~al.(2017)Solnik, Golovin, Kochanski, Karro, Moitra, and
  Sculley]{Solnik2017}
Benjamin Solnik, Daniel Golovin, Greg Kochanski, John~Elliot Karro, Subhodeep
  Moitra, and D.~Sculley.
\newblock Bayesian optimization for a better dessert.
\newblock In \emph{Proceedings of the 2017 NIPS Workshop on Bayesian
  Optimization}, December 9, 2017, Long Beach, USA, 2017.
\newblock The workshop is BayesOpt 2017 NIPS Workshop on Bayesian Optimization
  December 9, 2017, Long Beach, USA.

\bibitem[{Sun} et~al.(2016){Sun}, {Pan}, {White}, and {Dubey}]{Sun2016}
F.~{Sun}, Y.~{Pan}, J.~{White}, and A.~{Dubey}.
\newblock Real-time and predictive analytics for smart public transportation
  decision support system.
\newblock In \emph{2016 IEEE International Conference on Smart Computing
  (SMARTCOMP)}, May 2016.

\bibitem[Sun et~al.(2019)Sun, Jiang, Lam, Chen, and He]{Sun2019}
Yidan Sun, Guiyuan Jiang, Siew-Kei Lam, Shicheng Chen, and Peilan He.
\newblock Bus {Travel} {Speed} {Prediction} using {Attention} {Network} of
  {Heterogeneous} {Correlation} {Features}.
\newblock In \emph{Proceedings of ICDM}. Society for Industrial and Applied
  Mathematics, May 2019.
\newblock ISBN 978-1-61197-567-3.
\newblock URL \url{https://epubs.siam.org/doi/book/10.1137/1.9781611975673}.

\bibitem[{Transit App}(2015)]{transitapp}
{Transit App}.
\newblock "how we mapped the world’s weirdest streets", 2015.
\newblock URL
  \url{"https://medium.com/transit-app/hello-nairobi-cc27bb5a73b7"}.

\bibitem[{Transit Center}(2016)]{wob2016}
{Transit Center}.
\newblock Who's on board.
\newblock Technical report, {Transit Center}, 2016.
\newblock URL
  \url{http://transitcenter.org/wp-content/uploads/2016/07/Whos-On-Board-2016-7_12_2016.pdf}.

\bibitem[{Treethidtaphat} et~al.(2017){Treethidtaphat}, {Pattara-Atikom}, and
  {Khaimook}]{Treethidtaphat2017}
W.~{Treethidtaphat}, W.~{Pattara-Atikom}, and S.~{Khaimook}.
\newblock Bus arrival time prediction at any distance of bus route using deep
  neural network model.
\newblock In \emph{2017 IEEE 20th International Conference on Intelligent
  Transportation Systems (ITSC)}, pages 988--992, Oct 2017.

\bibitem[Vincent and Jerram(2006)]{Vincent2006}
William Vincent and Lisa~Callaghan Jerram.
\newblock The potential for bus rapid transit to reduce transportation-related
  $co_2$ emissions.
\newblock \emph{Journal of Public Transportation}, 9\penalty0 (3):\penalty0 12,
  2006.

\bibitem[Wan et~al.(2016)Wan, Liu, Shao, Vasilakos, Imran, and Zhou]{wan2016}
Jiafu Wan, Jianqi Liu, Zehui Shao, Athanasios~V. Vasilakos, Muhammad Imran, and
  Keliang Zhou.
\newblock Mobile crowd sensing for traffic prediction in internet of vehicles.
\newblock \emph{Sensors (Basel)}, 16\penalty0 (1), 2016.

\bibitem[Wang et~al.(2018)Wang, Zhang, Cao, Li, and Zheng]{Wang2018}
Dong Wang, Junbo Zhang, Wei Cao, Jian Li, and Yu~Zheng.
\newblock When will you arrive? {E}stimating travel time based on deep neural
  networks.
\newblock In \emph{Thirty-Second AAAI Conference on Artificial Intelligence},
  2018.

\bibitem[Watkins et~al.(2011)Watkins, Ferris, Borning, Rutherford, and
  Layton]{Watkins2011}
Kari~Edison Watkins, Brian Ferris, Alan Borning, G~Scott Rutherford, and David
  Layton.
\newblock Where is my bus? impact of mobile real-time information on the
  perceived and actual wait time of transit riders.
\newblock \emph{Transportation Research Part A: Policy and Practice},
  45\penalty0 (8):\penalty0 839--848, 2011.

\bibitem[Wessel et~al.(2017)Wessel, Allen, and Farber]{Wessel2017}
Nate Wessel, Jeff Allen, and Steven Farber.
\newblock Constructing a routable retrospective transit timetable from a
  real-time vehicle location feed and {GTFS}.
\newblock \emph{Journal of Transport Geography}, 62:\penalty0 92--97, 2017.

\bibitem[Xu and Ying(2017)]{Xu2017}
Haitao Xu and Jing Ying.
\newblock Bus arrival time prediction with real-time and historic data.
\newblock \emph{Cluster Computing}, 20\penalty0 (4):\penalty0 3099--3106,
  December 2017.
\newblock ISSN 1573-7543.

\bibitem[Zhang et~al.(2008)Zhang, Shen, and Clifton]{Zhang2008}
Feng Zhang, Qing Shen, and Kelly~J. Clifton.
\newblock Examination of traveler responses to real-time information about bus
  arrivals using panel data.
\newblock \emph{Transportation Research Record}, 2082\penalty0 (1):\penalty0
  107--115, 2008.

\bibitem[Zheng et~al.(2012)Zheng, Zhang, and Feng]{zheng2012improved}
Chang-Jiang Zheng, Yi-Hua Zhang, and Xue-Jun Feng.
\newblock Improved iterative prediction for multiple stop arrival time using a
  support vector machine.
\newblock \emph{Transport}, 27\penalty0 (2):\penalty0 158--164, 2012.

\end{thebibliography}
\bibliographystyle{plainnat}

\end{document}